\documentclass[runningheads]{llncs} 
\usepackage{named} 
\usepackage{times}
\usepackage{amsmath}
\usepackage{listings}
\usepackage{graphicx}

\long\def\BOC#1\EOC{\message{(Commented text )}}
\long\def\BOCC#1\EOCC{\message{(Commented text )}}
\long\def\BOCCC#1\EOCCC{\message{(Commented text )}}
\long\def\optional#1{\empty}

\def\ar{\leftarrow}
\def\bi{\begin{itemize}}
\def\ei{\end{itemize}}
\def\beq{\begin{equation}}
\def\eeq#1{\label{#1}\end{equation}}
\def\ba{\begin{array}}
\def\ea{\end{array}}
\def\i#1{\hbox{\it #1\/}}
\def\mi#1{\mathit{#1}}
\def\sm{\hbox{\rm SM}}

\def\sneg{\sim\!\!}
\def\ar{\leftarrow}
\def\rar{\rightarrow}
\def\lrar{\leftrightarrow}

\def\mvis{\!=\!}
\def\false{\hbox{\bf f}}
\def\true{\hbox{\bf t}}
\def\i#1{\hbox{\itshape #1\/}}

\def\mvis{\!=\!}

\def\caused{\hbox{\bf caused}}
\def\iif{\hbox{\bf if}}
\def\after{\hbox{\bf after}}
\def\ifcons{\hbox{\bf if}\;\!\hbox{\bf cons}}
\def\causes{\hbox{\bf causes}}
\def\inertial{\hbox{\bf inertial}}
\def\default{\hbox{\bf default}}
\def\constraint{\hbox{\bf constraint}}
\def\always{\hbox{\bf always}}

\def\nonexecutable{\hbox{\bf nonexecutable}}
\def\exogenous{\hbox{\bf exogenous}}

\def\pf{\hbox{\it PF}}

\newtheorem{thm}{Theorem}

\begin{document}

\mainmatter

\title{Action Language $\cal BC$+} 
\author{Joseph Babb \and Joohyung Lee\thanks{Corresponding author: joolee@asu.edu}}

\institute{
School of Computing, Informatics, and Decision Systems Engineering \\
Arizona State University, Tempe, USA}


\maketitle


\begin{abstract}
Action languages are formal models of parts of natural language that are designed to describe effects of actions. Many of these languages can be viewed as high level notations of answer set programs structured to represent transition systems. However, the form of answer set programs considered in the earlier work is quite limited in comparison with the modern Answer Set Programming (ASP) language, which allows several useful constructs for knowledge representation, such as choice rules, aggregates, and abstract constraint atoms. We propose a new action language called $\cal BC$+, which closes the gap between action languages and the modern ASP language. The main idea is to define the semantics of $\cal BC$+ in terms of general stable model semantics for propositional formulas, under which many modern ASP language constructs can be identified with shorthands for propositional formulas. Language $\cal BC$+ turns out to be sufficiently expressive to encompass the best features of other action languages, such as languages $\cal B$, $\cal C$, $\cal C$+, and $\cal BC$. Computational methods available in ASP solvers are readily applicable to compute $\cal BC$+, which led to an implementation of the language by extending system {\sc cplus2asp}.
\end{abstract}

\section{Introduction} \label{sec:intro}

Action languages are formal models of parts of natural language that are used for describing properties of actions. The semantics of action languages describe transition systems---directed graphs whose vertices represent states and whose edges represent actions that affect the states. Many action languages, such as languages $\cal A$ \cite{gel93a} and $\cal B$ \cite[Section~5]{gel98}, can be viewed as high level notations of answer set programs structured to represent transition systems.  
Languages $\cal C$ \cite{giu98} and $\cal C$+~\cite{giu04} are originally defined in terms of nonmonotonic causal theories, but their ``definite'' fragments can be equivalently turned into answer set programs as well~\cite{ferraris12representing}, which led to the implementation {\sc cplus2asp}, which uses ASP solvers for computation \cite{babb13cplus2asp}. 

The main advantage of using action languages over answer set programs is their structured abstract representations for describing transition systems, which allows their users to focus on high level descriptions and avoids the ``cryptic'' syntax and the recurring pattern of ASP rules for representing transition systems. However, existing work on action languages has two limitations. First, they do not allow several useful ASP language constructs, such as choice rules, aggregates, abstract constraint atoms, and external atoms, that have recently been introduced into ASP, and contributed to widespread use of ASP in many practical applications. The inability to express these modern constructs in action languages is what often forces the users to write directly in the language of ASP rather than in action languages.

Another issue arises even such constructs are not used: there are certain limitations that each action language has in comparison with one another. 
For instance, in language $\cal B$, the frame problem is solved by enforcing in the semantics that every fluent be governed by the commonsense law of inertia, which makes it difficult to represent fluents whose behavior is described by defaults other than inertia, such as the amount of water in a leaking container.
Languages $\cal C$ and $\cal C$+ do not have this limitation, but instead they do not handle Prolog-style recursive definitions available in $\cal B$. The recently proposed language $\cal BC$ \cite{lee13action} combines the attractive features of $\cal B$ and $\cal C$+, but it is not a proper generalization. In comparison with $\cal C$+, it does not allow us to describe complex dependencies among actions, thus it is unable to describe several concepts that $\cal C$+ is able to express, such as ``defeasible'' causal laws \cite[Section 4.3]{giu04} (causal laws that are retracted by adding additional causal laws) and ``attributes'' of an action~\cite[Section~5.6]{giu04}, which are useful for elaboration tolerant representation. 

We present a simple solution to these problems. The main idea is to define an action language in terms of a general stable model semantics, which has not been considered in the work on action languages.  We present a new action language called $\cal BC$+, which is defined as a high level notation of propositional formulas under the stable model semantics~\cite{fer05}. It has been well studied in ASP that several useful constructs, such as aggregates, abstract constraint atoms, and conditional literals, can be identified with abbreviations of propositional formulas (e.g.,\cite{fer05,pel03,sont07a,harrison14thesemantics}). 
Thus, $\cal BC$+ employs such constructs as well. Further, it is more expressive than the other action languages mentioned above, allowing them to be easily embedded.
Computational problems involving $\cal BC$+ descriptions can be reduced to computing answer sets. This fact led to an implementation of $\cal BC$+ by modifying system {\sc cplus2asp}~\cite{babb13cplus2asp}, which was originally designed to compute $\cal C$+ using ASP solvers. 

The paper is organized as follows. Section~\ref{sec:prelim} reviews propositional formulas under the stable model semantics for describing multi-valued constants.  Sections \ref{sec:syntax} and \ref{sec:semantics} present the syntax and the semantics of $\cal BC$+, and Section~\ref{sec:abbreviations} presents useful abbreviations of causal laws in that language, followed by Section~\ref{sec:examples}, which formalizes an example using such abbreviations. Section~\ref{sec:sm-bc+} shows how to embed propositional formulas under the stable model semantics into~$\cal BC$+.  
Sections~\ref{sec:bc} and \ref{sec:cplus} relate $\cal BC$+ to each of $\cal BC$ and $\cal C$+. Section~\ref{sec:implementation} describes an implementation of $\cal BC$+ as an extension of system {\sc cplus2asp}.
The proofs are given in Appendix~\ref{sec:proofs}.

This is an extended version of the conference paper \cite{babb15action}.

\section{Review: Propositional Formulas under the Stable Model Semantics} \label{sec:prelim}

A {\em propositional signature} is a set of symbols called {\em atoms}. A propositional formula is defined recursively using atoms and the following set of primitive propositional connectives: $\bot\hbox{ (falsity)},\ \land,\ \lor,\ \rar$.
We understand $\neg F$ as an abbreviation of $F\rar\bot$; symbol $\top$ stands for $\bot\rar\bot$, expression $G\ar F$ stands for $F\rar G$, and expression $F\lrar G$ stands for $(F\rar G)\land(G\rar F)$.

An {\em interpretation} of a propositional signature is a function from the signature into $\{\false,\true\}$. We identify an interpretation with the set of atoms that are true in it. 

A {\em model} of a formula is an interpretation that {\em satisfies} the formula. 
According to \cite{fer05}, the models are divided into stable models and non-stable models as follows.
The {\em reduct} $F^X$ of a propositional formula $F$ relative to a set $X$ of atoms is the formula obtained from $F$ by replacing every maximal subformula that is not satisfied by~$X$ with $\bot$. Set~$X$ is called a {\em stable model}  of $F$ if $X$ is a minimal set of atoms satisfying~$F^X$. 

Throughout this paper, we assume that the signature $\sigma$ is constructed from ``constants'' and their ``values.'' 
A {\em constant} $c$ is a symbol that is associated with a finite set $\i{Dom}(c)$ of cardinality $\ge 2$, called the {\em domain}. 
The signature $\sigma$ is constructed from a finite set of constants, consisting of atoms $c\!=\!v$~\footnote{%
Note that here ``='' is just a part of the symbol for propositional atoms, and is not  equality in first-order logic. }
for every constant $c$ and every element $v$ in $\i{Dom}(c)$.
If the domain of~$c$ is $\{\false,\true\}$ then we say that~$c$ is {\em Boolean}, and abbreviate $c\mvis\true$ as $c$ and $c\mvis\false$ as~$\sneg c$. 

\citeauthor{bartholomew14stable} [\citeyear{bartholomew14stable}] show that this form of propositional formulas is useful for expressing the concept of default values on multi-valued fluents. By $\i{UEC}_\sigma$ (``Uniqueness and Existence Constraint'') we denote the conjunction of 
\beq
  \bigwedge_{v\ne w\ :\  v,w \in\mi{Dom}(c)} \neg (c = v \land c =
  w),
\eeq{uc}
and
\beq
  \neg\neg\bigvee_{v \in \mi{Dom}(c)} c = v\ .
\eeq{ec}
for all constants $c$ of $\sigma$. 
It is clear that any propositional interpretation of $\sigma$ that satisfies $\i{UEC}_\sigma$ can be identified with a function that maps each constant $c$ into an element in its domain. 

\begin{example}\label{ex:1}
Consider a signature $\sigma$ to be $\{c\mvis 1,\ c\mvis 2,\ c\mvis 3\}$,
where $c$ is a constant and $\i{Dom}(c)=\{1,2,3\}$.
Formula~$\i{UEC}_\sigma$ is 
$$\neg (c\mvis 1\land c\mvis 2)\land \neg (c\mvis 2\land c\mvis 3)
\land\neg (c\mvis 1\land c\mvis 3) \land
\neg\neg (c\mvis 1\lor c\mvis 2\lor c\mvis 3). $$
Let $F_1$ be $(c\mvis 1\lor\neg (c\mvis 1))\land\i{UEC}_\sigma$. 
Due to $\i{UEC}_\sigma$, each of $\{c\mvis 1\}$, $\{c\mvis 2\}$, and $\{c\mvis 3\}$ is a model of $F_1$, but $\{c\mvis 1\}$ is the only stable model of $F_1$.
The reduct $F_1^{\{c=1\}}$ is equivalent to $c\mvis 1$, for which $\{c\mvis 1\}$ is the minimal model. On the other hand, for instance, the reduct $F_1^{\{c=2\}}$ is equivalent to $\top$, for which the minimal model is $\emptyset$, not $\{c\mvis 2\}$. 

Let $F_2$ be $F_1$ conjoined with $c\mvis 2$. Interpretation~$\{c\mvis 1\}$ is not a stable model of~$F_2$. Indeed, the reduct $F_2^{\{c=1\}}$ is $\bot$, for which there is no model. However, $\{c\mvis 2\}$ is a stable model of $F_2$. The reduct $F_2^{\{c=2\}}$ is equivalent to $c\mvis 2$, for which $\{c\mvis 2\}$ is the minimal model. 
This case illustrates the nonmonotonicity of the semantics. 
\end{example} 

Note that the presence of double negations is essential
in~\eqref{ec}. Without them, $F_1$ would have three stable models:
$\{c\mvis 1\}$, $\{c\mvis 2\}$, and $\{c\mvis 3\}$.

In ASP, formulas of the form $F\lor\neg F$ are called {\em choice formulas}, which we denote by $\{F\}^{\rm ch}$. For example, $F_1$ in Example~\ref{ex:1} can be written as $\{c\mvis 1\}^{\rm ch}\land\i{UEC}_\sigma$. 
As shown in Example~\ref{ex:1}, in the presence of $\i{UEC}_\sigma$, a formula of the form $\{c\mvis v\}^{\rm ch}$ expresses that $c$ has the value $v$ by default, which can be overridden in the presence of other evidences \cite{bartholomew14stable}.

Given that the domain is finite, aggregates in ASP can be understood as shorthand for propositional formulas as shown in \cite{fer05,pel03,sont07a,leej09a}. 
For instance, cardinality constraint (i.e., count aggregate)  $l\le Z$, where $l$ is a nonnegative integer, and $Z$ is a finite set of atoms, is the disjunction of the formulas $\bigwedge_{L\in Y} L$ over all $l$-element subset $Y$ of $Z$. 
For instance, the cardinality constraint $2\le\{p,q,r\}$ is shorthand for the propositional formula 
\[ 
  (p\land q)\lor (q\land r)\lor (p\land r). 
\]
Expression $Z\le u$, where $u$ is a nonnegative integer, denotes $\neg((u\!+\!1)\le Z)$. 
Expression $l\le Z\le u$ stands for $(l\le Z) \land (Z\le u)$.

More generally, abstract constraint atoms \cite{mare04} can be understood as shorthand for propositional formulas \cite{lee12stable}.

\section{Syntax of $\cal BC$+} \label{sec:syntax} 

The syntax of language $\cal BC$+ is similar to the syntax of $\cal C$+.\footnote{Strictly speaking, $\cal C$+ considers   ``multi-valued'' formulas, an extension of propositional formulas, but Theorem~1 from~\cite{bartholomew14stable} shows that multi-valued formulas under the stable model semantics can be identified with propositional formulas under the stable model semantics in the presence of the uniqueness and existence of value constraints.}
In language $\cal BC$+, a signature $\sigma$ is a finite set of propositional atoms of the form $c=v$, where constants $c$ are divided into two groups: {\em fluent constants} and {\em action constants}.  Fluent constants are further divided into {\em regular} and {\em statically determined}. 

A {\sl fluent formula} is a formula such that all constants occurring in it are fluent constants. An {\sl action formula} is a formula that contains at least one action constant and no fluent constants. \footnote{The definition implies that formulas that contain no constants (but may contain $\bot$ and $\top$) are fluent formulas.}

A {\em static law} is an expression of the form
\beq
  \caused\ F\ \iif\ G
\eeq{static}
where $F$ and $G$ are fluent formulas.  

An {\sl action dynamic law} is an expression of the form~(\ref{static}) in which $F$ is an action formula and $G$ is a formula.

A {\sl fluent dynamic law} is an expression of the form
\beq
 \caused\ F\ \iif\ G\ \after\ H
\eeq{dynamic}
where~$F$ and~$G$ are fluent formulas and $H$ is a formula, provided that~$F$ does not contain statically determined constants. 

Static laws can be used to talk about causal dependencies between fluents in the same state; action dynamic laws can be used to express causal dependencies between concurrently executed actions; fluent dynamic laws can be used to describe direct effects of actions. 

A {\sl causal law} is a static law, an action dynamic law, or a fluent dynamic law.
An {\sl action description} is a finite set of causal laws.

The formula~$F$ in causal laws~(\ref{static}) and~(\ref{dynamic}) is called the {\sl head}.

\section{Semantics  of $\cal BC$+}\label{sec:semantics}

For any action description~$D$ of a signature $\sigma$, we define a sequence of propositional formulas $\pf_0(D),\pf_1(D),\dots$ so that the stable models of $\pf_m(D)$ can be visualized as paths in a ``transition system''---a directed graph whose vertices are states of the world and edges represent transitions between states. 
The signature $\sigma_{m}$ of $\pf_m(D)$ consists of atoms of the form $i\!:\!c\mvis v$ such that
\begin{itemize}
\item  for each fluent constant $c$ of $D$, $i\in\{0,\dots,m\}$ and~$v\in\i{Dom}(c)$, and 
\item  for each action constant $c$ of $D$, $i\in\{0,\dots,m\!-\!1\}$ and~$v\in\i{Dom}(c)$. 
\end{itemize}
By $i\!:\!F$ we denote the result of inserting $i\!:$ in front of every occurrence of every constant in formula~$F$. This notation is similarly extended when $F$ is  a set of formulas. 
The translation~$\pf_m(D)$ is the conjunction of 
\bi
\item 
  \beq
    i\!:\!F\ \ar\  i\!:\!G 
  \eeq{static-pf}
  for every static law~(\ref{static}) in~$D$ and every~$i\in\{0,\dots,m\}$, and \eqref{static-pf} for every action dynamic law~(\ref{static}) in~$D$ and every~$i\in\{0,\dots,m\!-\!1\}$;

\item 
  \beq
    i\!+\!1\!:\!F\ \ar\ (i\!+\!1\!:\!G) \wedge (i\!:\!H)
  \eeq{dynamic-pf}
  for every fluent dynamic law~(\ref{dynamic}) in~$D$ and
  every~$i\in\{0,\dots,m\!-\!1\}$;

\item 
  \beq
    \{ 0\!:\!c\mvis v\}^{\rm ch} 
  \eeq{init-exog-pf}
  for every regular fluent constant~$c$ and every~$v\in\i{Dom}(c)$; 
\item 
  $\i{UEC}_{\sigma_{m}}$, which can also be abbreviated   
  using the count aggregate as 
\beq
    \bot\ar\neg\ (1\le \{i\!:\!c\mvis v_1, \dots, i\!:\!c\mvis v_m\}\le 1)
 \eeq{uec-pf}
where $\{v_1,\dots,v_m\}$ is $\i{Dom}(c)$. 
\ei

Note how the translation $\pf_m(D)$ treats regular and statically determined fluent constants differently: formulas~(\ref{init-exog-pf}) are included only when~$c$ is regular. Statically determined fluents are useful for describing defined fluents, whose values are determined by the fluents in the same state only. 
For instance, $\i{NotClear}(B)$ is a statically determined Boolean fluent constant defined by the static causal law
\[
\ba l
  \caused\ \i{NotClear}(B)\mvis\true\ \iif\  \i{Loc}(B_1)\mvis\ B,  \\
  \caused\ \{\i{NotClear}(B)\mvis\false\}^{\rm ch}. 
\ea
\]
If we added \eqref{init-exog-pf} for $\i{NotClear}(B)$, that is, 
\[
   \{0\!:\!\i{NotClear}(B)=\true\}^{\rm ch},\ \ 
   \{0\!:\!\i{NotClear}(B)=\false\}^{\rm ch},
\]
to the translation $\pf_m(D)$, the value of $\i{NotClear}(B)$ at time $0$ would have been arbitrary, which does not conform to the intended definition of $\i{NotClear}(B)$.
We refer the reader to \cite[Section~5]{giu04} for more details about the difference between regular and statically determined fluent constants.

\begin{figure}
\centering
\includegraphics[width=6cm]{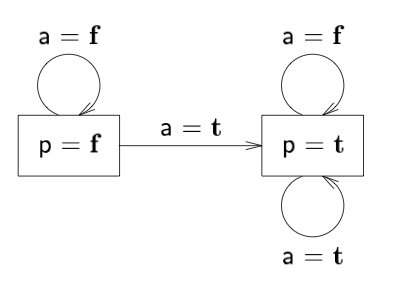}
\caption{The transition system described by~$\i{SD}$.}
\label{fig:sd}
\end{figure}

\begin{example}
The transition system shown in Figure~\ref{fig:sd} can be described by the following action description $\i{SD}$, where $p$ is a Boolean regular fluent constant and $a$ is a Boolean action constant. 
\beq
\ba l
  \caused\ p\ \iif\ \top\ \after\ a, \\
  \caused\ \{a\}^{\rm ch}\ \iif\ \top, \\
  \caused\ \{\sneg a\}^{\rm ch}\ \iif\ \top, \\
  \caused\ \{p\}^{\rm ch}\ \iif\ \top\ \after\ p,  \\
  \caused\ \{\sneg p\}^{\rm ch}\ \iif\ \top\ \after\ \sneg p.
\ea
\eeq{sd}

The translation $\pf_m(\i{SD})$ turns this description into the
following propositional formulas. 
The first line of~(\ref{sd}) is turned into the formulas (disregarding $\top$)
$$
  i\!+\!1\!:\!p\ \ar\ i\!:\!a  
$$
($0\le i< m$), the second and the third lines into
\beq
\ba l
  \{ i\!:\!a\}^{\rm ch}, \hspace{1cm} \\
  \{ i\!:\sneg a\}^{\rm ch}
\ea
\eeq{s1}
($0\le i< m$), and the fourth and the fifth lines into
\beq
\ba l
  \{ i\!+\!1\!:\!p\}^{\rm ch}\ \ar\ i\!:\!p, \hspace{1cm} \\
  \{ i\!+\!1\!:\sneg p\}^{\rm ch}\ \ar\ i\!:\sneg p\ 
\ea
\eeq{s2}
$(0\le i< m)$.
In addition, 
$$
\ba l
\{ 0\!:\!p\}^{\rm ch}, \hspace{1cm} \\
\{ 0\!:\sneg p\}^{\rm ch}
\ea
$$
come from~(\ref{init-exog-pf}), and 
\[
\ba l
  \bot\ar \neg\ (1\le\{i\!:\!p,\ i\!:\sneg p\}\le 1) \qquad (0\le i\le m),\\
  \bot\ar \neg\ (1\le\{i\!:\!a,\ i\!:\sneg a\}\le 1) \qquad (0\le i< m)
\ea
\]
come from \eqref{uec-pf}.
\end{example}

Let $\sigma^{fl}$ be the subset of the signature $\sigma$ consisting of atoms containing fluent constants, and let $\sigma^{act}$ be the subset of $\sigma$ consisting of atoms containing action constants.
Since we identify an interpretation $I$ with the set of atoms that are true in it, an interpretation of the signature $\sigma_{m}$ can be represented in the form
\[
  (0:s_0)\cup (0:e_0) \cup (1:s_1) \cup (1:e_1)\cup \dots\cup (m:s_m)
\]
where $s_0,\dots,s_m$ are interpretations of $\sigma^{fl}$, and $e_0,\dots,e_{m-1}$ are interpretations of~$\sigma^{act}$.

We define states and transitions in terms of stable models of $\pf_0(D)$ and $\pf_1(D)$ as follows.

\begin{definition}[States and Transitions]\label{def:state-trans}
For any action description~$D$ of signature $\sigma$, a {\em state} of $D$ is an interpretation~$s$ of~$\sigma^{fl}$ such that $0\!:\!s$ is a stable model of~$\pf_0(D)$. 
A {\em transition} of $D$ is a triple $\langle s,e,s'\rangle$ where $s$ and $s'$ are interpretations of~$\sigma^{fl}$ and~$e$ is an interpretation of~$\sigma^{act}$ such that $0\!:\!s \cup 0\!:\!e \cup 1\!:\!s'$ is a stable model of $\pf_1(D)$.
\end{definition} 

In view of the uniqueness and existence of value constraints for every state~$s$ and every fluent constant~$c$, there exists exactly one~$v$ such that~$c\!=\!v$ belongs to~$s$; this~$v$ is considered the value of~$c$ in state~$s$.

Given these definitions, we define the transition system $T(D)$ represented by an action description $D$ as follows. 

\begin{definition}[Transition System]\label{def:transition-system}
The {\em transition system} $T(D)$ represented by an action description $D$ is a labeled directed graph such that the vertices are the states of $D$, and the edges are obtained from the transitions of $D$:  for every transition $\langle s,e,s'\rangle$ of $D$, an edge labeled $e$ goes from $s$ to~$s'$. 
\end{definition}

Since the vertices and the edges of a transition system $T(D)$ are identified with the states and the transitions of $D$, we simply extend the definitions of a state and a transition to transition systems: A {\em state} of $T(D)$ is a state of $D$. A {\em transition} of $T(D)$ is a transition of $D$.

The soundness of this definition is guaranteed by the following fact:

\begin{thm} \label{thm:state}
For every transition $\langle s, e, s'\rangle$ of~$D$, interpretations $s$ and $s'$ are states of~$D$.
\end{thm}

The stable models of $\pf_m(D)$ represent the paths of length~$m$ in the transition system represented by~$D$.  For $m=0$ and $m=1$, this is clear from the definition of a transition system (Definition~\ref{def:transition-system}); for $m>1$ this needs to be verified as the following theorem shows.

For every set~$X_m$ of elements of the signature~$\sigma_{m}$, let $X^i$ ($i<m$) be the triple consisting of
\begin{itemize}
\item  the set consisting of atoms~$A$ such that $i\!:\!A$ belongs to~$X_m$, and $A$
  contains fluent constants,
\item  the set consisting of atoms~$A$ such that $i\!:\!A$ belongs to~$X_m$, and $A$
  contains action constants, and
\item  the set consisting of atoms~$A$ such that $(i+1)\!:\!A$ belongs to~$X_m$, and
  $A$ contains fluent constants.
\end{itemize}

\begin{thm}\label{thm:trans}\optional{thm:trans}
For every $m\geq 1$,
$X_m$ is a stable model of~$\pf_m(D)$
iff $X^0,\dots,X^{m-1}$ are transitions of~$D$.
\end{thm}

For example, $\{0\!:\sneg p, 0\!:\sneg a, 1\!:\sneg p, 1\!:\!a, 2\!:\! p\}$ is a stable model of $\pf_2(\i{SD})$, and each of
$\langle\{\sneg p\}, \{\sneg a\}, \{\sneg p\}\rangle$ and 
$\langle\{\sneg p\}, \{a\}, \{p\}\rangle$ is a transition of $\i{SD}$.

\section{Useful Abbreviations} \label{sec:abbreviations}

Like $\cal C$+, several intuitive abbreviations of causal laws can be
defined for $\cal BC$+.

Expression 
\[ 
  \default\ c\mvis v\ \iif\ F  
\]
stands for 
\[
  \caused\ \{ c\mvis v\}^{\rm ch}\ \iif\ F .\footnote{%
Here and after, we often omit $\iif\ F$ if $F$ is $\top$.}
\]
This abbreviation is intuitive in view of the reading of choice formulas in the presence of the uniqueness and existence of value constraints (recall Example~\ref{ex:1}).  Similarly, 
\[
   \default\ c\mvis v\ \iif\ F\ \after\ G
\]
stands for 
\[
   \caused\ \{ c\mvis v\}^{\rm ch}\ \iif\ F\ \after\ G. 
\]

Other abbreviations of $\cal BC$+ causal laws are defined similarly to abbreviations in $\cal C$+. 
\bi 
\item If~$c$ is a Boolean action constant, we express
that $F$ is an effect of executing $c$ by 
\[ 
  c\ \causes\ F,
\]
which stands for the fluent dynamic law 
\[
   \caused\ F\ \iif\ \top\ \after\ c.
\]

\item If~$c$ is an action constant, the expression
\[ 
  \exogenous\ c
\]
stands for the action dynamic laws
\[ 
  \default\  c\mvis v
\]
for all $v\in\i{Dom}(c)$.

\item If~$c$ is a regular fluent constant, the expression
\[
  \inertial\ c
\]
stands for the fluent dynamic laws
\[ 
   \default\  c\mvis v\ \after\ c\mvis v
\]
for all $v\in\i{Dom}(c)$.  

\item $$\constraint\ F$$ where $F$ is a fluent formula stands for the
  static law 
\[ 
     \caused\ \bot\ \iif\ \neg F.
\]

\item $$\always\ F$$
stands for the fluent dynamic law
\[ 
    \caused\ \bot\ \iif\ \top\ \after\ \neg F.
\]

\item
\[ 
   \nonexecutable\ F\ \iif\ G 
\]
stands for the fluent dynamic law
\[
  \caused\ \bot\ \iif\ \top\ \after\ F\land G.
\]
\ei

\section{Example: Blocks World} \label{sec:examples}

An attractive feature of $\cal BC$+ is that aggregates are directly usable in causal laws because they can be understood as abbreviations of propositional formulas~\cite{fer05,pel03,sont07a,leej09a}.
We illustrate this advantage by formalizing an elaboration of the Blocks World from~\cite{lee13action}.

Let $\i{Blocks}$ be a nonempty finite set $\{\i{Block}_1, \dots, \i{Block}_n\}$.
The action description below uses the following fluent and action constants:
\begin{itemize}
\item  for each $B\in\i{Blocks}$, regular fluent constant $\i{Loc}(B)$
  with the domain \hbox{$\i{Blocks}\cup\{\i{Table}\}$}, and statically
  determined Boolean fluent constant $\i{InTower}(B)$;

\item for each $B\in\i{Blocks}$, Boolean action constant $\i{Move}(B)$;

\item for each $B\in\i{Blocks}$, action constant 
  $\i{Destination}(B)$ with the domain
  $\i{Blocks}\cup\{\i{Table}\}\cup\{\i{None}\}$, where $\i{None}$ is a
  symbol for denoting an ``undefined'' value. 
\end{itemize}
In the list of static and dynamic laws, $B$, $B_1$ and $B_2$ are arbitrary
elements of~\i{Blocks}, and~$L$ is an arbitrary element of
$\i{Blocks}\cup\{\i{Table}\}$.  Below  we list causal laws describing this domain. 

Blocks are not on itself:
\[
   \constraint\ \i{Loc}(B)\!\ne\! B. 
\]

The definition of $\i{InTower}(B)$:
\beq
\ba l
  \caused\ \i{InTower}(B)\ \iif\ \i{Loc}(B)\mvis\i{Table},\\
  \caused\ \i{InTower}(B)\ \iif\ \i{Loc}(B)\mvis B_1\land \i{InTower}(B_1),\\
  \default\ \sneg\i{InTower}(B).
\ea
\eeq{intower}

Blocks do not float in the air:
\[
  \constraint\ \i{InTower}(B).
\]

No two blocks are on the same block: 
\[
  \constraint\ \{b: \i{Loc}(b)\mvis B\}\le 1, 
\]
which is shorthand for
\[
  \constraint\ \{\i{Loc}(\i{Block}_1)\mvis B, \dots, \i{Loc}(\i{Block}_n)\mvis B\}\le 1.
\]

Only ${\rm k}$ towers are allowed to be on the table (${\rm k}$ is a positive integer): 
\[
  \constraint\ \{b: \i{Loc}(b)\mvis\i{Table}\}\le {\rm k}.
\]

The effect of moving a block:
\[ 
   \i{Move}(B)\ \causes\ \i{Loc}(B)\mvis L\ \iif\ \i{Destination}(B)\mvis L. 
\]

A block cannot be moved unless it is clear:
\[ 
   \nonexecutable\ \i{Move}(B)\ \iif\ \i{Loc}(B_1)\mvis B.
\]

Concurrent actions are limited by the number ${\rm g}$ of grippers:
\[
  \always\ \{b: \i{Move}(b)\}\le {\rm g}. 
\]

The commonsense law of inertia:
\[ 
  \inertial\ \i{Loc}(B).
\]

Actions are exogenous: 
\[
\ba l
  \exogenous\ \i{Move}(B),\\
  \exogenous\ \i{Destination}(B). 
\ea
\]

$\i{Destination}$ is an attribute of $\i{Move}$: 
\[
  \always\ \i{Destination}(B)\mvis \i{None}\lrar\neg\i{Move}(B).
\]

Besides the inability to represent aggregates, other action languages have other difficulties in representing this example. 
Under the semantics of $\cal C$ and $\cal C$+, the recursive definition of $\i{InTower}$ in~\eqref{intower} does not work correctly.
Languages $\cal B$ and $\cal BC$ do not allow us to represent action attributes like $\i{Destination}$ because they lack nonBoolean actions and action dynamic laws (The usefulness of attributes in expressing elaboration tolerance was discussed in~\cite{lif00}.)

\section{Embedding Formulas under SM in  $\cal BC$+} \label{sec:sm-bc+}

We defined the semantics of~$\cal BC$+ by reducing the language to propositional formulas under the stable model semantics. The reduction in the opposite direction is also possible.

For any propositional formula $F$, we define the translation $\mathsf{pf2bcp}(F)$, which turns $F$ into an ``equivalent'' action description in $\cal BC$+ as follows: reclassify every atom in the signature of~$F$ as a statically determined fluent constant with Boolean values, and rewrite $F$ as the static law
\[ 
  \caused\ F
\]
and add 
\[
  \default\ c\mvis\false
\] 
for every constant $c$. 

We identify an interpretation $I$ of the signature of $F$ with an interpretation $I'$ of the signature of $\mathsf{pf2bcp}(D)$ as follows: for all atoms $c$ in the signature of $F$, 
$I(c)=v$ iff $I'\models c\mvis v$ ($v\in\{\true,\false\}$). Due to the presence of $\i{UEC}_{\{c\mvis{\bf t}, c\mvis{\bf f}\}}$ in $\mathsf{pf2bcp}(D)$, the mapping also tells us that any interpretation satisfying $\mathsf{pf2bcp}(D)$ has a corresponding interpretation of the signature of~$F$.

\begin{thm}\label{thm:pf2bcp}\optional{thm:pf2bcp}
For any propositional formula $F$ of a finite signature and any interpretation $I$ of that signature,  $I$ is a stable model of $F$ iff $I'$ is a state of the transition system represented by the $\cal BC$+ description $\mathsf{pf2bcp}(F)$.
\end{thm}

It is known that the problem of determining the existence of stable models of propositional formulas is $\Sigma_2^P$-complete \cite{fer05}. The same complexity applies to $\cal BC$+ in view of Proposition~\ref{thm:pf2bcp}. On the other hand, from the translation $\i{PF}_m(D)$, a useful fragment in NP can be defined based on the known results in ASP.  The following is an instance, which we call ``simple'' action descriptions.

We say that action description $D$ is {\em definite} if the head of every causal law is either $\bot$, an atom $c\mvis v$, or a choice formula  $\{c\mvis v\}^{\rm ch}$. 
We say that a formula is a {\em simple} conjunction if it is a conjunction of atoms and count aggregate expressions, each of which possibly preceded by negation.  
A {\sl simple} action description is a definite action description such that $G$ in every causal law~\eqref{static} is a simple conjunction, and $G$ and $H$ in every causal law~\eqref{dynamic} are simple conjunctions. 
The Blocks World formalization in the previous section is an example of a simple action description.

\section{Relation to Language $\cal BC$} \label{sec:bc}

\subsection{Review: $\cal BC$}

The signature $\sigma$ for a $\cal BC$ description $D$ is defined the same as in $\cal BC$+ except that every action constant is assumed to be Boolean-valued. The main syntactic differences between $\cal BC$ causal laws and $\cal BC$+ causal laws are that the former allows only the conjunction of atoms in the body, and distinguishes between $\iif$ and $\hbox{\bf if\;\!cons}$ clauses. 

A {\em $\cal BC$ static law} is an expression of the form
\beq
   A_0\ \iif\ A_1,\dots,A_m\ \ifcons\ A_{m+1},\dots,A_n
\eeq{static-bc}
$(n\geq m\geq 0)$ where each~$A_i$ is an atom containing a fluent constant.  It expresses, informally speaking, that every state satisfies~$A_0$ if it satisfies $A_1,\dots,A_m$, and $A_{m+1},\dots,A_n$ can be consistently assumed. 

A {\em $\cal BC$ dynamic law} is an expression of the form
\beq
  A_0\ \after\ A_1,\dots,A_m\ \ifcons\ A_{m+1},\dots,A_n
\eeq{dynamic-bc}
$(n\geq m\geq 0)$ where 
\begin{itemize}
\item  $A_0$ is an atom containing a regular fluent constant,
\item  each of $A_1,\dots,A_m$ is an atom containing a fluent constant, or $a\mvis\true$ where $a$ is an action constant, and
\item  $A_{m+1},\dots,A_n$ are atoms containing fluent constants.
\end{itemize}
It expresses, informally speaking, that the end state of any transition satisfies~$A_0$ if its beginning state and its action satisfy $A_1,\dots,A_m$, and $A_{m+1},\dots,A_n$ can be consistently assumed about the end state. 

An {\em action description} in language $\cal BC$ is a finite set of $\cal BC$ static and $\cal BC$ dynamic laws.

Like $\cal BC$+, the semantics of $\cal BC$ is defined by reduction $\pf_m^{\cal BC}$ to a sequence of logic programs under the stable model semantics. 
The signature $\sigma_{m}$ of $\pf_m^{\cal BC}$ is defined the same as that of $\pf_m$ defined in Section~\ref{sec:semantics}.

For any $\cal BC$ action description~$D$, by~$\pf_m^{\cal BC}(D)$ we denote the conjunction of
\begin{itemize}
\item 
\beq
i\!:\!A_0\ \ar\ i:(A_1\land\cdots\land A_m\land\neg\neg A_{m+1}\land\dots\land\neg\neg A_n)
\eeq{static-bc-pf}
for every $\cal BC$ static law~\eqref{static-bc} in $D$ and every $i\in\{0,\dots,m\}$;

\item 
\beq
\ba l
(i+1)\!:\!A_0\ \ar\ 
i\!:\!(A_1\land\cdots\land A_m)\land 
(i\!+\!1)\!:\!(\neg\neg A_{m+1}\land\dots\land\neg\neg A_n)
\ea
\eeq{dynamic-bc-pf}
for every $\cal BC$ dynamic law~\eqref{dynamic-bc} in $D$ and every $i\in\{0,\dots,m\!-\!1\}$;

\item
the formula $i\!:\!(a\!=\!\true\lor a\!=\!\false)$ for every action constant~$a$ and every \hbox{$i\in\{0,\dots,m\!-\!1\}$};

\item
the formula \eqref{init-exog-pf} for every regular fluent constant~$c$ and every element~$v\in\i{Dom}(c)$;

\item $\i{UEC}_{\sigma_{m}}$.
\end{itemize}

Note how the translations~\eqref{static-bc-pf} and \eqref{dynamic-bc-pf} treat $\iif$ and $\hbox{{\bf ifcons}}$ clauses differently by either prepending double negations in front of atoms or not. 
In $\cal BC$+, only one $\iif$ clause is enough since the formulas are understood under the stable model semantics. We explore this difference in more detail in Section~\ref{ssec:bc+-bc}.

\subsection{Embedding $\cal BC$ in $\cal BC$+}

Despite the syntactic differences, language $\cal BC$ can be easily embedded in $\cal BC$+ as follows. 
For any $\cal BC$ description $D$, we define the translation $\mathsf{bc2bcp}(D)$, which turns a $\cal BC$ description into an equivalent $\cal BC$+ description as follows:
\bi
\item replace every causal law \eqref{static-bc} with
  \[ 
     \caused\ A_0\ \iif\ 
        A_1\land\dots\land A_m\land\neg\neg A_{m+1}\land\dots\land\neg\neg
        A_n; 
  \] 
\item replace every causal law \eqref{dynamic-bc} with
  \[
     \caused\ A_0\ \iif\ \neg\neg A_{m+1}\land\dots\land\neg\neg A_n\ 
       \after\ A_1\land\dots\land A_m; 
  \]
\item add the causal laws
\[
  \exogenous\ a
\]
 for every action constant $a$. 
\ei

\begin{thm}\label{thm:bc2bcp}\optional{thm:bc2bcp}
For any action description $D$ in language $\cal BC$, the transition system described by $D$ is identical to the transition system described by the description $\mathsf{bc2bcp}(D)$ in language $\cal BC$+.
\end{thm}

\subsection{Comparing $\cal BC$+ with $\cal BC$} \label{ssec:bc+-bc}

In $\cal BC$, every action is assumed to be Boolean, and action dynamic laws are not available, which prevents us from describing defeasible causal laws \cite[Section 4.3]{giu04} and action attributes \cite[Section~5.6]{giu04}, that $\cal BC$+ and $\cal C$+ are able to express conveniently. Also, syntactically, $\cal BC$ is not expressive enough to describe dependencies among actions. For a simple example, in $\cal BC$+ and $\cal C$+, we can express that action $a_1$ is not executable when $a_2$ is not executed at the same time by the fluent dynamic law 
\[
   \caused\ \bot\ \after\ a_1\land\neg a_2, 
\]
but this is not even syntactically allowed in $\cal BC$.

On the other hand, the presence of choice formulas in the head of $\cal BC$+ causal laws  and the different treatment of $A$ and $\neg\neg A$ in the bodies may look subtle to those who are not familiar with the stable model semantics for propositional formulas. Fortunately, in many cases the subtlety can be avoided by using the $\default$ proposition (Section~\ref{sec:abbreviations})  as the following example illustrates.

Consider the leaking container example from~\cite{lee13action} in which a container loses $k$ units of liquid by default. This example was used to illustrate the advantages of $\cal BC$ over $\cal B$ that is able to express defaults other than inertia. In this domain, the default decrease of $\i{Amount}$ over time can be represented in $\cal BC$+ using the $\default$ abbreviation 
\beq
  \default\ \i{Amount}\mvis x\ \after\ \i{Amount}\mvis x\!+\!k, 
\eeq{amount}
which stands for fluent dynamic law
\[
  \caused\ \{\i{Amount}\mvis x\}^{\rm ch}\ \after\ \i{Amount}\mvis x\!+\!k, 
\]
which is shorthand for propositional formulas
\beq
   \{i\!+\!1\!:\!\i{Amount}\mvis x\}^{\rm ch} \ar i\!:\!\i{Amount}\mvis x\!+\!k
\eeq{amount-1}
($i<m$).

The $\default$ abbreviation is also defined in $\cal BC$ in a syntactically different, but semantically equivalent way. In $\cal BC$, the assertion \eqref{amount} stands
for the causal law
\[
  \caused\ \i{Amount}\mvis x\ \after\ \i{Amount}\mvis x\!+\!k\ \ifcons\
  \i{Amount}\mvis x, 
\]
which is further turned into
\[
   i\!+\!1\!:\!\i{Amount}\mvis x\ \ar i\!:\!\i{Amount}\mvis x\!+\!k \land
           \neg\neg (i\!+\!1\!:\!\i{Amount}\mvis x)
\]
($i<m$), 
which is strongly equivalent to \eqref{amount-1}.\footnote{
About propositional formulas~$F$ and~$G$ we say that~$F$ is {\sl strongly equivalent} to~$G$ if, for every propositional formula~$H$, $F\land H$ has the same stable models as~$G\land H$ \cite{lif01}.}


\section{Relation to $\cal C$+} \label{sec:cplus}

\subsection{Review: $\cal C$+}\label{ssec:review-cplus}

As mentioned earlier, the syntax of $\cal C$+ is similar to the syntax of $\cal BC$+. The signature is defined the same as in $\cal BC$+. A {\em $\cal C$+ static law} is an expression of the form~\eqref{static} where $F$ and $G$ are fluent formulas. A {\em $\cal C$+ action dynamic law} is an expression of the form~\eqref{static} in which $F$ is an action formula and $G$ is a formula. A {\sl $\cal C$+ fluent dynamic law} is an expression of the form~\eqref{dynamic} where~$F$ and~$G$ are fluent formulas and $H$ is a formula, provided that~$F$ does not contain statically determined constants. A {\sl $\cal C$+ causal law} is a static law, an action dynamic law, or a fluent dynamic law. A {\sl $\cal C$+ action description} is a set of $\cal C$+ causal laws. We say that $\cal C$+ action description $D$ is {\em definite} if the head of every causal law is either $\bot$ or an atom $c\mvis v$.

The original semantics of $\cal C$+ is defined in terms of reduction to nonmonotonic causal theories in \cite{giu04}. In~\cite{lee12reformulating2}, the semantics of the definite $\cal C$+ description is equivalently reformulated in terms of reduction to propositional formulas under the the stable model semantics as follows. \footnote{The translation does not work for nondefinite $\cal C$+ descriptions, due to the different treatments of the heads under nonmonotonic causal theories and under the stable model semantics.}

For any definite $\cal C$+ action description~$D$ and any nonnegative
integer~$m$, the propositional formula $\pf_m^{{\cal C}+}(D)$ is defined as
follows. The signature of $\pf_m^{{\cal C}+}(D)$ is defined the same
as $\pf_m(D)$.
The translation~$\pf_m^{{\cal C}+}(D)$ is the conjunction of
\bi
\item 
\beq
  i\!:\!F \ar\ \neg\neg\, (i\!:\!G)\
\eeq{tr1-fsm}
for every static law~(\ref{static}) in~$D$ and every~$i\in\{0,\dots,m\}$,
and for every action dynamic law~(\ref{static}) in~$D$ and
every~$i\in\{0,\dots,m\!-\!1\}$;

\item 
\beq
 i\!+\!1\!:\!F\ar\ \neg\neg (i\!+\!1\!:\!G) \land (i\!:\!H) \ 
\eeq{tr2-fsm}
for every fluent dynamic law~(\ref{dynamic})
in~$D$ and every~$i\in\{0,\dots,m\!-\!1\}$;

\item the formula \eqref{init-exog-pf} for every regular fluent
  constant~$c$ and every~$v\in\i{Dom}(c)$;

\item $\i{UEC}_{\sigma_{m}}$.
\ei

\subsection{Embedding Definite $\cal C$+ in $\cal BC$+}

For any definite $\cal C$+ description $D$, we define the
translation $\mathsf{cp2bcp}(D)$, which turns a $\cal C$+ description into
$\cal BC$+, as follows:

\bi
\item replace every $\cal C$+ causal law \eqref{static} with
  \[
     \caused\ F\ \iif\ \neg\neg G; 
  \]
\item replace every $\cal C$+ causal law \eqref{dynamic} with
  \[
     \caused\ F\ \iif\ \neg\neg G\
       \after\ H.
  \]
\ei

The following theorem asserts the correctness of this translation. 

\begin{thm}\label{thm:cp2bcp}\optional{thm:cp2bcp}
For any definite action description $D$ in language $\cal C$+, the transition system described by $D$ is identical to the transition system described by the description $\mathsf{cp2bcp}(D)$ in language $\cal BC$+.
\end{thm}


\subsection{Comparing $\cal BC$+ with $\cal C$+} \label{ssec:adv-cplus}

\begin{figure}[t!]
\hrule
\smallskip
\begin{tabbing}
Notation:  $s,s'$ range over $\{\i{Switch}_1, \i{Switch}_2\}$;\ \ \
 $x, y$ range over $\{\i{Up}, \i{Down}\}$.\\ 

Regular fluent constants:          \hskip 2cm  \=Domains:\\
$\;$ $\i{Status}(s)$                 \>$\;$ $\{\i{Up}, \i{Down}\}$\\
Action constants:                          \>Domains:\\
$\;$ $\i{Flip}(s)$                    \>$\;$ Boolean\\ \\
Causal laws:\\ \\
$\i{Flip}(s)\ \causes\ \i{Status}(s)\mvis x\ \iif\ \i{Status}(s)\mvis
  y \hspace{1cm} (x\ne y)$ \\
$\caused\ \i{Status}(s)\mvis x\ \iif\ \i{Status}(s')\mvis y
\hspace{1cm} (s\ne s', x\ne y)$ \\
$\inertial\ \i{Status}(s)$ \\
$\exogenous\ \i{Flip}(s)$
\end{tabbing}
\hrule
\caption{Two Switches in $\cal BC$+} 
\label{fig:switch}
\end{figure}

The embedding of $\cal C$+ in $\cal BC$+ tells us that ${\bf if}$ clauses always introduce double negations, whose presence leads to the fact that stable models are not necessarily minimal models. This accounts for the fact that the definite fragment of $\cal C$+ does not handle the concept of transitive closure correctly.  For example, the recursive definition of $\i{InTower}(B)$ in Section~\ref{sec:examples} does not work correctly if it is understood as $\cal C$+ causal laws.
The inability to consider minimal models in such cases introduces some unintuitive behavior of $\cal C$+ in representing causal dependencies among fluents, as the following example shows.



%
Consider two switches which can be flipped but cannot be both up or
down at the same time.\footnote{This example is similar to Lin's suitcase example \cite{lin95}, but a main difference is that the same fluent is affected by both direct and indirect effects of an action.}
If one of them is down and the other is up, the direct effect of flipping only one switch is changing the status of that switch, and the indirect effect is changing the status of the other switch. 
This domain can be represented in $\cal BC$+ as shown in Figure~\ref{fig:switch}. 

The description in $\cal BC$+ has the following four transitions
possible from the initial state where $\i{Switch}_1$ is $\i{Down}$ and
$\i{Switch}_2$ is $\i{Up}$:

{\footnotesize
\noindent
\begin{itemize}
\item 
$
\ba l
\langle \{\i{St}(\i{Sw}_1)\mvis\i{Dn}, \i{St}(\i{Sw}_2)\mvis\i{Up}
\},  \{ \sneg\i{Flip}(\i{Sw}_1), \sneg\i{Flip}(\i{Sw}_2) \},
\{\i{St}(\i{Sw}_1)\mvis\i{Dn},  \i{St}(\i{Sw}_2)\mvis\i{Up} \}\rangle,
\ea
$

\noindent 
\item $
\ba l
\langle \{\i{St}(\i{Sw}_1)\mvis\i{Dn}, \i{St}(\i{Sw}_2)\mvis\i{Up} \}, 
        \{ \i{Flip}(\i{Sw}_1), \sneg\i{Flip}(\i{Sw}_2) \}, 
\{\i{St}(\i{Sw}_1)\mvis\i{Up}, 
         \i{St}(\i{Sw}_2)\mvis\i{Dn} \}\rangle,
\ea
$

\noindent
\item $
\ba l
\langle \{\i{St}(\i{Sw}_1)\mvis\i{Dn}, \i{St}(\i{Sw}_2)\mvis\i{Up} \}, 
        \{ \sneg\i{Flip}(\i{Sw}_1), \i{Flip}(\i{Sw}_2) \}, 
        \{\i{St}(\i{Sw}_1)\mvis\i{Up}, \i{St}(\i{Sw}_2)\mvis\i{Dn} \}\rangle,
\ea 
$

\noindent
\item $
\ba l
\langle \{\i{St}(\i{Sw}_1)\mvis\i{Dn}, \i{St}(\i{Sw}_2)\mvis\i{Up} \}, 
         \{ \i{Flip}(\i{Sw}_1), \i{Flip}(\i{Sw}_2) \}, 
     \{\i{St}(\i{Sw}_1)\mvis\i{Up}, 
         \i{St}(\i{Sw}_2)\mvis\i{Dn} \}\rangle.
\ea
$
\end{itemize}
}

The second and the third transitions exhibit the indirect effect of the action $\i{Flip}$.
If this description is understood in $\cal C$+, five transitions are possible from the same initial state: in addition to the four transitions above, 

{\footnotesize
\noindent
\begin{itemize}
\item 
$
\ba l
\langle \{\i{St}(\i{Sw}_1)\mvis\i{Dn}, \i{St}(\i{Sw}_2)\mvis\i{Up} \}, 
         \{ \sneg\i{Flip}(\i{Sw}_1), \sneg\i{Flip}(\i{Sw}_2) \},
\{\i{St}(\i{Sw}_1)\mvis\i{Up}, 
         \i{St}(\i{Sw}_2)\mvis\i{Dn} \}\rangle
\ea 
$
\end{itemize}
}
\noindent
is also a transition because, according to the semantics of $\cal C$+, this is causally explained by the cyclic causality. This is obviously unintuitive.

\section{Implementation} \label{sec:implementation}

System {\sc cplus2asp} \cite{babb13cplus2asp} was originally designed to compute the definite fragment of $\cal C$+ using ASP solvers as described in \cite{lee13action}. 
Its version~2 supports extensible multi-modal translations for other action languages as well. As the translation $\pf_m^{\cal C+}(D)$ for $\cal C$+ is similar to the translation $\pf_m(D)$ for $\cal BC$+, the extension is straightforward. We modified system {\sc cplus2asp} to be able to accept $\cal BC$+ as another input language. 

The $\cal BC$+ formalization of the Blocks World domain can be represented in the input language of {\sc cplus2asp} under the $\cal BC$+ mode as shown in Figure~\ref{fig:blocks-cplus2asp}.

\begin{figure}
\hrule\smallskip
\begin{lstlisting}
% File 'blocks'

:- sorts
    location >> block.

:- objects
    b(1..10)                :: block;
    table                   :: location.

:- constants
    loc(block)	            :: inertialFluent(location);
    in_tower(block)         :: sdFluent;
    move(block)             :: exogenousAction;
    dest(block)             :: attribute(location*) of move(block).

:- variables
    B,B1,B2                 :: block;
    L	                    :: location.

constraint -(loc(B)=B).

caused in_tower(B) if loc(B)=table.
caused in_tower(B) if loc(B)=B1 & in_tower(B1).
default ~in_tower(B).

constraint in_tower(B).
constraint {B1| loc(B1)=B}1.
constraint {B1| loc(B1)=table}k.
move(B) causes loc(B)=L if dest(B)=L.
nonexecutable move(B) if loc(B1)=B.
always {B1| move(B1)}g.

:- query
label :: test;
0: loc(b(1))=table & loc(b(2))=b(1) & loc(b(3))=b(2) 
   & loc(b(4))=b(3) & loc(b(5))=b(4);
0: loc(b(6))=table & loc(b(7))=b(6) & loc(b(8))=b(7) 
   & loc(b(9))=b(8) & loc(b(10))=b(9);
maxstep: loc(b(1))=b(10).
\end{lstlisting}
\hrule
\caption{Blocks World in the input language of {\sc cplus2asp} under the $\cal BC$+ mode}
\label{fig:blocks-cplus2asp}
\end{figure}

The input language of {\sc cplus2asp} allows the users to conveniently express declarations and causal laws. Its syntax follows the syntax of Version 2 of the {\sc CCalc} input language.\footnote{%
\url{http://www.cs.utexas.edu/users/tag/ccalc/}
}
The extent of each sort (i.e., domain) is defined in the object declaration section. The sort declaration denotes that {\tt block} is a subsort of {\tt location}, meaning that every object of sort {\tt block} is an object of {\tt location} as well. 
The constant declaration
\begin{lstlisting}
  loc(block)             :: inertialFluent(location)
\end{lstlisting}
has the same meaning as the declaration
\begin{lstlisting}
  loc(block)             :: simpleFluent(location)
\end{lstlisting}
({\tt simpleFluent} is a keyword for regular fluent) 
accompanied by the dynamic law
\begin{lstlisting}
  inertial loc(B).
\end{lstlisting}
The constant declaration
\begin{lstlisting}
  destination(block)    :: attribute(location*) of move(block).
\end{lstlisting}
has the same meaning as the declaration
\begin{lstlisting}
  destination(block)    :: action(location*) 
\end{lstlisting}
accompanied by the dynamic law
\begin{lstlisting}
  exogenous destination(B).
  always destination(B)=none <-> -move(B).
\end{lstlisting}
where {\tt location*} is a new sort implicitly declared by {\sc cplus2asp} and consists of {\tt location}s and the auxiliary symbol {\tt none}.

We refer the reader to the system homepage 
\begin{center}
  {\tt http://reasoning.eas.asu.edu/cplus2asp}
\end{center}
for the details of the input language syntax. In order to run this program we invoke {\sc cplus2asp} as follows.
\begin{lstlisting}
   cplus2asp -l bc+ blocks k=3 g=2 query=test
\end{lstlisting}
The option {\tt -l bc+} instructs {\sc cplus2asp} to operate under the $\cal BC$+ semantics. ``{\tt k=3 g=2}'' are constant definitions for the number {\tt k} of towers, and the number {\tt g} of grippers for the domain.

\begin{figure}[h!]
\hrule\smallskip
\begin{lstlisting}
% File 'switch'

:- sorts
   switch; status.

:- objects
   s1, s2             :: switch;
   on, off            :: status.

:- constants
   sw_status(switch)  :: inertialFluent(status);
   flip(switch)       :: exogenousAction.

:- variables
   S, S1              :: switch;
   X, Y               :: status.

flip(S) causes sw_status(S)=X if sw_status(S)=Y & X\=Y.

caused sw_status(S)=X if sw_status(S1)=Y & S\=S1 & X\=Y.

:- query
   label :: test;
   maxstep :: 1;
   0: sw_status(s1)=off & sw_status(s2)=on.
\end{lstlisting}
\hrule
\caption{Two Switches in the input language of {\sc cplus2asp} under the $\cal BC$+ mode}
\label{fig:switch-cplus2asp}
\end{figure}

For another example, Figure~\ref{fig:switch-cplus2asp} represents the $\cal BC$+ description in Figure~\ref{fig:switch} in the input language of {\sc cplus2asp}.
The following is the command line to find all four transitions described in Section~\ref{ssec:adv-cplus}. 
\begin{lstlisting}
  $ cplus2asp -l bc+ switch query=test 0 
\end{lstlisting}

The $0$ at the end instructs the system to find all stable models. 
The following is the output: 

\begin{lstlisting}
Solution: 1
	0:  sw_status(s1)=off sw_status(s2)=on

	1:  sw_status(s1)=off sw_status(s2)=on

Solution: 2
	0:  sw_status(s1)=off sw_status(s2)=on

	ACTIONS:  flip(s1) flip(s2)

	1:  sw_status(s1)=on sw_status(s2)=off

Solution: 3
	0:  sw_status(s1)=off sw_status(s2)=on

	ACTIONS:  flip(s2)

	1:  sw_status(s1)=on sw_status(s2)=off

Solution: 4
	0:  sw_status(s1)=off sw_status(s2)=on

	ACTIONS:  flip(s1)

	1:  sw_status(s1)=on sw_status(s2)=off
\end{lstlisting}

If the same program is run under the $\cal C$+ mode, 
\begin{lstlisting}
  cplus2asp -l c+ switch query=test 0 
\end{lstlisting}
one more (unintuitive) transition is returned: 
\begin{lstlisting}
	0:  sw_status(s1)=off sw_status(s2)=on

	1:  sw_status(s1)=on sw_status(s2)=off
\end{lstlisting}

\section{Conclusion} 

Unlike many other action languages which can be understood as high level notations of limited forms of logic programs, $\cal BC$+ is defined as a high level notation of the general stable model semantics for propositional formulas. This approach allows for employing modern ASP language constructs directly in $\cal BC$+, as they can be understood as shorthand for propositional formulas, and thus allows for closing the gap between action languages and the modern ASP language. It also accounts for the expressivity of $\cal BC$+ for embedding other action languages, and allows reasoning about transition systems described in $\cal BC$+ to be computed by ASP solvers. 


A further extension of $\cal BC$+ is possible by replacing the role of propositional formulas with a more expressive generalized stable model semantics. It is straightforward to extend $\cal BC$+ to the first-order level by using the first-order stable model semantics from~\cite{ferraris11stable} or its extension with generalized quantifiers~\cite{lee12stable} in place of propositional formulas. 
This will allow $\cal BC$+ to include other constructs, such as external atoms and nonmonotonic dl-atoms, as they are instances of generalized quantifiers as shown in~\cite{lee12stable}.

Also, some recent advances in ASP solving can be applied to action languages. Our future work includes extending $\cal BC$+ to handle external events arriving online based on the concept of online answer set solving~\cite{gebser11reactive}, and compute it using  online answer set solver {\sc oclingo}.\footnote{%
\url{http://www.cs.uni-potsdam.de/wv/oclingo/}}

\section*{Acknowledgements}

We are grateful to Michael Bartholomew, Vladimir Lifschitz and the anonymous referees for their useful comments. 
This work was partially supported by the National Science Foundation under Grant IIS-1319794, South Korea IT R\&D program MKE/KIAT 2010-TD-300404-001, and 
ICT R\&D program of MSIP/IITP 10044494 (WiseKB).

\bibliographystyle{named}

\begin{thebibliography}{}

\bibitem[\protect\citeauthoryear{Babb and Lee}{2013}]{babb13cplus2asp}
Joseph Babb and Joohyung Lee.
\newblock {C}plus2{A}{S}{P}: Computing action language {$\cal C$+} in answer
  set programming.
\newblock In {\em Proceedings of International Conference on Logic Programming
  and Nonmonotonic Reasoning (LPNMR)}, pages 122--134, 2013.

\bibitem[\protect\citeauthoryear{Babb and Lee}{2015}]{babb15action}
Joseph Babb and Joohyung Lee.
\newblock Action language {$\cal BC$}+: Preliminary report.
\newblock In {\em Proceedings of the AAAI Conference on Artificial Intelligence
  (AAAI)}, 2015.

\bibitem[\protect\citeauthoryear{Bartholomew and
  Lee}{2012}]{bartholomew12stable}
Michael Bartholomew and Joohyung Lee.
\newblock Stable models of formulas with intensional functions.
\newblock In {\em Proceedings of International Conference on Principles of
  Knowledge Representation and Reasoning (KR)}, pages 2--12, 2012.

\bibitem[\protect\citeauthoryear{Bartholomew and
  Lee}{2014}]{bartholomew14stable}
Michael Bartholomew and Joohyung Lee.
\newblock Stable models of multi-valued formulas: partial vs. total functions.
\newblock In {\em Proceedings of International Conference on Principles of
  Knowledge Representation and Reasoning (KR)}, pages 583--586, 2014.

\bibitem[\protect\citeauthoryear{Ferraris \bgroup \em et al.\egroup
  }{2009}]{ferraris09symmetric}
Paolo Ferraris, Joohyung Lee, Vladimir Lifschitz, and Ravi Palla.
\newblock Symmetric splitting in the general theory of stable models.
\newblock In {\em Proceedings of International Joint Conference on Artificial
  Intelligence (IJCAI)}, pages 797--803. AAAI Press, 2009.

\bibitem[\protect\citeauthoryear{Ferraris \bgroup \em et al.\egroup
  }{2011}]{ferraris11stable}
Paolo Ferraris, Joohyung Lee, and Vladimir Lifschitz.
\newblock Stable models and circumscription.
\newblock {\em Artificial Intelligence}, 175:236--263, 2011.

\bibitem[\protect\citeauthoryear{Ferraris \bgroup \em et al.\egroup
  }{2012}]{ferraris12representing}
Paolo Ferraris, Joohyung Lee, Yuliya Lierler, Vladimir Lifschitz, and Fangkai
  Yang.
\newblock Representing first-order causal theories by logic programs.
\newblock {\em TPLP}, 12(3):383--412, 2012.

\bibitem[\protect\citeauthoryear{Ferraris}{2005}]{fer05}
Paolo Ferraris.
\newblock Answer sets for propositional theories.
\newblock In {\em Proceedings of International Conference on Logic Programming
  and Nonmonotonic Reasoning ({LPNMR})}, pages 119--131, 2005.

\bibitem[\protect\citeauthoryear{Gebser \bgroup \em et al.\egroup
  }{2011}]{gebser11reactive}
Martin Gebser, Torsten Grote, Roland Kaminski, and Torsten Schaub.
\newblock Reactive answer set programming.
\newblock In {\em Proceedings of International Conference on Logic Programming
  and Nonmonotonic Reasoning (LPNMR)}, pages 54--66. Springer, 2011.

\bibitem[\protect\citeauthoryear{Gelfond and Lifschitz}{1993}]{gel93a}
Michael Gelfond and Vladimir Lifschitz.
\newblock Representing action and change by logic programs.
\newblock {\em Journal of Logic Programming}, 17:301--322, 1993.

\bibitem[\protect\citeauthoryear{Gelfond and Lifschitz}{1998}]{gel98}
Michael Gelfond and Vladimir Lifschitz.
\newblock Action languages. 
\newblock {\em Electronic Transactions on Artificial Intelligence}, 3:195--210,
  1998.
\newblock {\tt http://www.ep.liu.se/ea/cis/1998/016/}.

\bibitem[\protect\citeauthoryear{Giunchiglia and Lifschitz}{1998}]{giu98}
Enrico Giunchiglia and Vladimir Lifschitz.
\newblock An action language based on causal explanation: Preliminary report.
\newblock In {\em Proceedings of National Conference on Artificial Intelligence
  ({AAAI})}, pages 623--630. AAAI Press, 1998.

\bibitem[\protect\citeauthoryear{Giunchiglia \bgroup \em et al.\egroup
  }{2004}]{giu04}
Enrico Giunchiglia, Joohyung Lee, Vladimir Lifschitz, Norman McCain, and Hudson
  Turner.
\newblock Nonmonotonic causal theories.
\newblock {\em Artificial Intelligence}, 153(1--2):49--104, 2004.

\bibitem[\protect\citeauthoryear{Harrison \bgroup \em et al.\egroup
  }{2014}]{harrison14thesemantics}
Amelia~J. Harrison, Vladimir Lifschitz, and Fangkai Yang.
\newblock The semantics of Gringo and infinitary propositional formulas.
\newblock In {\em Principles of Knowledge Representation and Reasoning:
  Proceedings of the Fourteenth International Conference, {KR} 2014}, 2014.

\bibitem[\protect\citeauthoryear{Lee and Meng}{2009}]{leej09a}
Joohyung Lee and Yunsong Meng.
\newblock On reductive semantics of aggregates in answer set programming.
\newblock In {\em Proceedings of International Conference on Logic Programming
  and Nonmonotonic Reasoning (LPNMR)}, pages 182--195, 2009.

\bibitem[\protect\citeauthoryear{Lee and Meng}{2012}]{lee12stable}
Joohyung Lee and Yunsong Meng.
\newblock Stable models of formulas with generalized quantifiers.
\newblock In {\em Proceedings of International Workshop on Nonmonotonic
  Reasoning (NMR)}, 2012.
\newblock {\tt http://peace.eas.asu.edu/joolee/papers/smgq-nmr.pdf}.

\bibitem[\protect\citeauthoryear{Lee \bgroup \em et al.\egroup
  }{2013}]{lee13action}
Joohyung Lee, Vladimir Lifschitz, and Fangkai Yang.
\newblock Action language {$\cal BC$}: Preliminary report.
\newblock In {\em Proceedings of International Joint Conference on Artificial
  Intelligence (IJCAI)}, 2013.

\bibitem[\protect\citeauthoryear{Lee}{2012}]{lee12reformulating2}
Joohyung Lee.
\newblock Reformulating action language {$\cal C$}+ in answer set programming.
\newblock In Esra Erdem, Joohyung Lee, Yuliya Lierler, and David Pearce,
  editors, {\em Correct Reasoning}, volume 7265 of {\em Lecture Notes in
  Computer Science}, pages 405--421. Springer, 2012.

\bibitem[\protect\citeauthoryear{Lifschitz \bgroup \em et al.\egroup
  }{2001}]{lif01}
Vladimir Lifschitz, David Pearce, and Agustin Valverde.
\newblock Strongly equivalent logic programs.
\newblock {\em ACM Transactions on Computational Logic}, 2:526--541, 2001.

\bibitem[\protect\citeauthoryear{Lifschitz}{2000}]{lif00}
Vladimir Lifschitz.
\newblock Missionaries and cannibals in the {C}ausal {C}alculator.
\newblock In {\em Proceedings of International Conference on Principles of
  Knowledge Representation and Reasoning (KR)}, pages 85--96, 2000.

\bibitem[\protect\citeauthoryear{Lin}{1995}]{lin95}
Fangzhen Lin.
\newblock Embracing causality in specifying the indirect effects of actions.
\newblock In {\em Proceedings of International Joint Conference on Artificial
  Intelligence ({IJCAI})}, pages 1985--1991, 1995.

\bibitem[\protect\citeauthoryear{Marek and Truszczynski}{2004}]{mare04}
Victor~W. Marek and Miroslaw Truszczynski.
\newblock Logic programs with abstract constraint atoms.
\newblock In {\em Proceedings of National Conference on Artificial Intelligence
  ({AAAI})}, pages 86--91, 2004.

\bibitem[\protect\citeauthoryear{Pelov \bgroup \em et al.\egroup
  }{2003}]{pel03}
Nikolay Pelov, Marc Denecker, and Maurice Bruynooghe.
\newblock Translation of aggregate programs to normal logic programs.
\newblock In {\em Proceedings Answer Set Programming}, 2003.

\bibitem[\protect\citeauthoryear{Son and Pontelli}{2007}]{sont07a}
Tran~Cao Son and Enrico Pontelli.
\newblock A constructive semantic characterization of aggregates in answer set
  programming.
\newblock {\em TPLP}, 7(3):355--375, 2007.

\end{thebibliography}

\appendix

\section{Proofs of Theorems~\ref{thm:state} and \ref{thm:trans}} \label{sec:proofs}

For the proofs below, it is convenient to use the following generalization over the stable model semantics from~\cite{fer05}, which is the propositional case of the first-order stable model semantics from~\cite{ferraris11stable}.

For any two interpretations $I$, $J$ of the same propositional signature and any list ${\bf p}$ of distinct atoms, we write $J<^{\bf p} I$ if
\begin{itemize}
\item  $J$ and $I$ agree on all atoms not in ${\bf p}$, and 

\item  $J$ is a proper subset of $I$. 
\end{itemize}

$I$ is a stable model of $F$ relative to ${\bf p}$, denoted by $I\models \sm[F; {\bf p}]$,  if $I$ is a model of~$F$ and there is no interpretation $J$ such that $J <^{\bf p} I$ and $J$ satisfies $F^I$.

When ${\bf p}$ is empty, this notion of a stable model coincides with the notion of a model in propositional logic. When ${\bf p}$ is the same as the underlying signature, the notion reduces to the notion of a stable model from \cite{fer05}. 

\bigskip\noindent
{\bf Theorem~\ref{thm:state}}\optional{thm:state}\ \ 
{\sl 
For every transition $\langle s, e, s'\rangle$ of~$D$, $s$ and $s'$ are states of~$D$.
}
\medskip

\begin{proof}
We will use the following notations: 
$\i{SD}(i)$ is the set of formulas~\eqref{static-pf} in~$\pf_m(D)$
obtained from the static laws \eqref{static} in $D$;
$\i{AD}(i)$ is the set of formulas~\eqref{static-pf} in~$\pf_m(D)$
obtained from the action dynamic laws \eqref{static} in $D$;
$\i{FD}(i)$ is the set of formulas~\eqref{dynamic-pf} in~$\pf_m(D)$
obtained from the fluent dynamic laws \eqref{dynamic} in~$D$.
For the signature $\sigma$ of $D$, signature~$\sigma^{r}$ is the subset of $\sigma$ consisting of atoms containing regular fluent constants; signature~$\sigma^{sd}$ is the subset of $\sigma$ consisting of atoms containing statically determined fluent constants; signature~$\sigma^{fl}$ is the union of $\sigma^{r}$ and $\sigma^{sd}$; signature~$\sigma^{act}$ is the subset of $\sigma$ consisting of atoms containing action constants.

Since $\langle s, e, s'\rangle$ is a transition, 
\[
\ba l
  0\!:\!s\cup 0\!:\!e\cup 1\!:\!s'\models \\
~~~~~~~~   \sm[\i{SD}(0)\cup\i{AD}(0)\cup\i{FD}(0)\cup\i{SD}(1);\  0\!:\!\sigma^{sd}\cup 0\!:\!\sigma^{act}\cup 
          1\!:\!\sigma^{r}\cup
          1\!:\!\sigma^{sd}]\\ 
\hspace{2em}\ \land\ \i{UEC}_{0:\sigma^{fl}\cup
            0:\sigma^{act}\cup 1:\sigma^{fl}}. 
\ea 
\]
By the splitting theorem \cite{ferraris09symmetric}, it follows that
\beq
  0\!:\!s\cup 0\!:\!e\cup 1\!:\!s'\models \sm[\i{SD}(0);\  0\!:\!\sigma^{sd}];
\eeq{state-1} 
\beq
  0\!:\!s\cup 0\!:\!e\cup 1\!:\!s'\models \sm[\i{FD}(0)\land\i{SD}(1);\ 
          1\!:\!\sigma^{r}\cup 1\!:\!\sigma^{sd}];
\eeq{state-2}
 \[
   0\!:\!s\cup 0\!:\!e\cup 1\!:\!s'\models 
      \i{UEC}_{0:\sigma^{fl}\cup 0:\sigma^{act}\cup 1:\sigma^{fl}}.
 \]

From~\eqref{state-1}, we have $0\!:\!s\models\sm[\i{SD}(0);\ 0\!:\!\sigma^{sd}]$, and consequently,  \\
$0\!:\!s\models\sm[\i{SD}(0)\land\i{UEC}_{0:\sigma^{fl}};\  0\!:\!\sigma^{sd}]$,
so $s$ is a state. 

From~\eqref{state-2}, by Theorem~2 from~\cite{ferraris11stable}, it
follows that
\beq
  0\!:\!s\cup 0\!:\!e\cup 1\!:\!s'\models \sm[\i{FD}(0)\land\i{SD}(1);\ 
          1\!:\!\sigma^{sd}].
\eeq{state-3}
Since $\i{FD}(0)$ is negative on $1\!:\!\sigma^{sd}$ (cf. \cite{ferraris11stable}),  \eqref{state-3} is
equivalent to 
\[
  0\!:\!s\cup 0\!:\!e\cup 1\!:\!s'\models \sm[\i{SD}(1);\ 
          1\!:\!\sigma^{sd}]\land \i{FD}(0), 
\]
Consequently, we get
\[
  1\!:\!s'\models \sm[\i{SD}(1)\land\i{UEC}_{1:\sigma^{fl}};\ 1\!:\!\sigma^{sd}],
\]
which can be rewritten as 
\[
  0\!:\!s'\models \sm[\i{SD}(0)\land\i{UEC}_{0:\sigma^{fl}};\ 0\!:\!\sigma^{sd}],
\]
so $s'$ is a state.
\end{proof}

\bigskip
\noindent
{\bf Theorem~\ref{thm:trans}}\optional{thm:trans}\ \ 
{\sl 
For every $m\geq 1$,
$X_m$ is a stable model of~$\pf_m(D)$
iff $X^0,\dots,X^{m-1}$ are transitions of~$D$.
}\medskip

\begin{proof}
When $m=1$, the claim is immediate from the definition of a
transition. 

I.H. Assume that $X_m$ is a stable model of $\pf_m(D)$ iff 
$X^0,\dots,X^{m-1}$ are transitions of $D$~($m\ge 1$).

We first prove that if $X_{m+1}$ is a stable model of $\pf_{m+1}(D)$, then $X^0,\dots, X^{m+1}$ are transitions of $D$.  Assume that $X_{m+1}$ is a stable model of $\pf_{m+1}(D)$.

Note that $X_{m+1}$ is a stable model of $\pf_{m+1}(D)$ iff $X_{m+1}$ satisfies
$\i{UEC}_{\sigma_{m+1}}$ and 
\beq
\ba l
    \sm[\i{SD}(0)\land\i{AD}(0)\land\i{FD}(0)\land\i{SD}(1) \\ 
    \hspace{4.5em} \land\ \i{AD}(1)\land\i{FD}(1)\land\i{SD}(2) \\
    \hspace{4.5em} \land\ \dots \\
    \hspace{4.5em} \land\ \i{AD}(m\!-\!1)\land\i{FD}(m\!-\!1)\land\i{SD}(m) \\
    \hspace{4.5em} \land\ \i{AD}(m)\land\i{FD}(m)\land\i{SD}(m\!+\!1)\ ; \\

\hspace{14em} 0\!:\!\sigma^{sd}\cup 0\!:\!\sigma^{act}\cup 
    1\!:\!\sigma^{r}\cup 1\!:\!\sigma^{sd}  \\ 
\hspace{16.7em} \cup\ 1\!:\!\sigma^{act}\cup 2\!:\!\sigma^{r}\cup 2\!:\!\sigma^{sd} \\
\hspace{16.7em} \cup\dots \\
\hspace{16.7em} \cup\ (m\!-\!1)\!:\!\sigma^{act}\cup m\!:\!\sigma^{r}\cup m\!:\!\sigma^{sd} \\
\hspace{16.7em}\cup\ m\!:\!\sigma^{act}\cup (m\!+\!1)\!:\!\sigma^{r}\land
    (m\!+\!1)\!:\!\sigma^{sd}].
\ea
\eeq{trans-0}
By the splitting theorem \cite{ferraris09symmetric}, the fact that $X_{m+1}$
satisfies \eqref{trans-0} is equivalent to saying that $X_{m+1}$ satisfies
\beq
\ba l
    \sm[\i{SD}(0)\land\i{AD}(0)\land\i{FD}(0)\land\i{SD}(1) \\ 
    \hspace{4.5em} \land\ \i{AD}(1)\land\i{FD}(1)\land\i{SD}(2) \\
    \hspace{4.5em} \land\ \dots \\
    \hspace{4.5em} \land\
    \i{AD}(m\!-\!1)\land\i{FD}(m\!-\!1)\land\i{SD}(m)\ ;
    \\
\hspace{14em} 0\!:\!\sigma^{sd}\cup 0\!:\!\sigma^{act}\cup 
    1\!:\!\sigma^{r}\cup 1\!:\!\sigma^{sd}\\
\hspace{16.7em} \cup\ 1\!:\!\sigma^{act}\cup 2\!:\!\sigma^{r}\cup 2\!:\!\sigma^{sd}\\
\hspace{16.7em} \cup\dots \\
\hspace{16.7em} \cup\ (m\!-\!1)\!:\!\sigma^{act}\cup m\!:\!\sigma^{r}\cup m\!:\!\sigma^{sd}]
\ea
\eeq{trans-1}
and 
\beq
\ba l
   \sm[\i{AD}(m)\land\i{FD}(m)\land\i{SD}(m\!+\!1);\  
       m\!:\!\sigma^{act}\cup (m\!+\!1)\!:\!\sigma^r\cup (m\!+\!1)\!:\!\sigma^{sd}]. 
\ea
\eeq{trans-2}

The fact that $X_{m+1}\models\eqref{trans-1}$ is equivalent to saying that 
\[ 
  X_{m+1}\cap\{0\!:\!\sigma\cup\dots\cup m\!:\!\sigma\}\models\eqref{trans-1}.
\]
By I.H., the latter is equivalent to saying that $X^0,\dots,X^{m-1}$ are transitions of $D$.

Observe that, using the splitting theorem, \eqref{trans-1} entails 
\beq
  \sm[\i{FD}(m\!-\!1)\land\i{SD}(m);\ 
                m\!:\!\sigma^r\cup m\!:\!\sigma^{sd}].
\eeq{trans-3}
By Theorem~2 from~\cite{ferraris11stable}, \eqref{trans-3} entails
\beq
    \sm[\i{FD}(m\!-\!1)\land\i{SD}(m);\ m\!:\!\sigma^{sd}]. 
\eeq{trans-4}
Since $\i{FD}(m\!-\!1)$ is negative on $m\!:\!\sigma^{sd}$,
\eqref{trans-4} entails
\beq
  \sm[\i{SD}(m);\ m\!:\!\sigma^{sd}]. 
\eeq{trans-5}
By the splitting theorem on \eqref{trans-2} and \eqref{trans-5}, $X_{m+1}$ satisfies
\beq
\ba l
     \sm[\i{SD}(m)\land\i{AD}(m)\land\i{FD}(m)\land\i{SD}(m\!+\!1);\  \\
\hspace{8em}  m\!:\!\sigma^{sd}\cup m\!:\!\sigma^{act}\cup 
             (m\!+\!1)\!:\!\sigma^r\cup
             (m\!+\!1)\!:\!\sigma^{sd}], 
\ea
\eeq{trans-6}
or equivalently, 
\[ 
\ba l
   0\!:\!X^m\models 
     \sm[\i{SD}(0)\land\i{AD}(0)\land\i{FD}(0)\land\i{SD}(1);\  0\!:\!\sigma^{sd}\cup 0\!:\!\sigma^{act}\cup 1\!:\!\sigma^r\cup
         1\!:\!\sigma^{sd}]. 
\ea
\]
The latter, together with the fact that $0\!:\!X^m$ satisfies $\i{UEC}_{\sigma_1}$, means that $X^m$ is a transition of~$D$.

\bigskip
We next prove that if $X^0,\dots, X^{m+1}$ are transitions of $D$, then $X_{m+1}$ is a stable model of $\pf_{m+1}(D)$. Assume that $X^0,\dots, X^{m+1}$ are transitions of $D$. By I.H., it follows that $X_{m+1}$ satisfies \eqref{trans-1} and~\eqref{trans-6}. 
By the splitting theorem on \eqref{trans-6}, $X_{m+1}$ satisfies 
\beq
  \sm[\i{AD}(m)\land\i{FD}(m)\land\i{SD}(m\!+\!1);\  
             m\!:\!\sigma^{act}\cup 
             (m\!+\!1)\!:\!\sigma^r\cup
             (m\!+\!1)\!:\!\sigma^{sd}].
\eeq{trans-7} 
From the fact that $X_{m+1}$ satisfies \eqref{trans-1} and \eqref{trans-7}, by the splitting theorem, we get $X_{m+1}$ satisfies \eqref{trans-0}. It is clear that $X_{m+1}$ satisfies $\i{UEC}_{\sigma_{m+1}}$. Consequently, $X_{m+1}$ is a stable model of~$\pf_m(D)$.
\end{proof}

\bigskip
\noindent
{\bf Theorem~\ref{thm:pf2bcp}}\ \ 
{\sl 
For any propositional formula $F$ of a finite signature and any interpretation $I$ of that signature,  $I$ is a stable model of $F$ iff $I'$ is a state of the transition system represented by the $\cal BC$+ description $\mathsf{pf2bcp}(F)$.
}
\medskip

\begin{proof}
We refer the reader to \cite{bartholomew14stable} for the definition of a multi-valued formula. 

Let ${\bf c}$ be the propositional signature of $F$. We identify ${\bf c}$ with the multi-valued signature where each atom is identified with a Boolean constant, and identify $F$ with the multi-valued formula in the multi-valued signature by identifying every occurrence of an atom $c$ with $c\mvis\true$. Further, we identify an interpretation of the propositional signature~${\bf c}$ with an interpretation of the multi-valued signature as follows: a propositional interpretation $I$ satisfies a propositional atom $c$ iff $I$, understood as a multi-valued interpretation, satisfies the multi-valued atom $c\mvis\true$.

By Corollary~1 (a) from~\cite{bartholomew12stable}, 
$$
\text{$I$ is a propositional stable model of $F$ relative to ${\bf c}$}
$$
is equivalent to saying that 
\beq
\text{$I$ is a multi-valued stable model of $F\land\i{DF}$ relative to ${\bf c}$},
\eeq{mv-f-df}
where $\i{DF}$ is the conjunction of $\{c\mvis \false\}^{\rm ch}$ for all $c\in {\bf c}$.

Let ${\bf c}'$ be the propositional signature consisting of $c\mvis\true, c\mvis\false$ for all $c$ in the multi-valued signature ${\bf c}$, and let $I'$ be the interpretation of ${\bf c}'$ such that $I(c)=v$ iff $I'\models c\mvis v$ ($v\in\{\true,\false\}$).

By Theorem~1 from~\cite{bartholomew14stable}, \eqref{mv-f-df} is equivalent to saying that 
$$
\text{
$I'$ is a propositional stable model of $F\land\i{DF}\land\i{UEC}_{{\bf c}'}$ relative to ${\bf c}'$.}
$$ 
The conclusion follows because $0: (F\land\i{DF}\land\i{UEC}_{{\bf c}'})$ is the formula resulting from $\mathsf{pf2bcp}(F)$.
\end{proof}

\bigskip
\noindent
{\bf Theorem~\ref{thm:bc2bcp}\optional{thm:bc2bcp}}\ \ 
{\sl 
For any action description $D$ in language $\cal BC$, the transition system described by $D$ is identical to the transition system described by the description $\mathsf{bc2bcp}(D)$ in language $\cal BC$+.
}
\medskip

\begin{proof}
The proof can be established by showing strong equivalence between the propositional formula $\pf_m^{\cal BC}(D)$ and the propositional formula $\pf_m(\mathsf{bc2bcp(D)})$.
The only non-trivial thing to check is that, for each action $a$, $$(a=\true\lor a=\false)\land\i{UEC}_{\{a\mvis{\bf t}, a\mvis{\bf f}\}}$$ is strongly equivalent to 
$\{a\mvis\true\}^{\rm ch}\land\{a\mvis\false\}^{\rm ch}\land\i{UEC}_{\{a\mvis{\bf t}, a\mvis{\bf f}\}}$. 
In view of Theorem~9 from~\cite{ferraris11stable}, it is sufficient to check that 
under the assumption that \hbox{$(a^*\rar a)\land\i{UEC}_{\{a\mvis{\bf t}, a\mvis{\bf f}\}}$},
\[
   (a^*\mvis\true\lor a^*\mvis\false)
\]
is classically equivalent to 
\[ 
   (a^*\mvis\true\lor \neg(a\mvis\true)) \land (a^*\mvis\false\lor \neg(a\mvis\false)), 
\]
which is clear.
\end{proof}

\bigskip
\noindent
{\bf Theorem~\ref{thm:cp2bcp}\optional{thm:cp2bcp}}\ \ \ 
{\sl
For any definite action description $D$ in language $\cal C$+, the
transition system described by $D$ is identical to the transition
system described by the description $\mathsf{cp2bcp}(D)$ in language $\cal BC$+. 
}
\medskip

\begin{proof}
The proof can be established by showing the strong equivalence between the propositional formula $\pf_m^{{\cal C}+}(D)$ and the propositional formula $\pf_m(\mathsf{cp2bcp}(D))$. This is easy to check.
\end{proof}

\end{document}